\documentclass[a4paper]{article}

\usepackage{float}
\usepackage{amsmath}
\usepackage{txfonts}
\usepackage{color}
\usepackage{graphicx}
\usepackage{booktabs}
\usepackage{multirow}
\usepackage{array}
\usepackage{orcidlink}
\usepackage{authblk}
\usepackage{url}

\makeatletter
\def\NOLTA@banner{}
\def\NOLTA@license{}
\makeatother

\title{\textbf{Sign Language Recognition using Bidirectional Reservoir Computing}}

\author[1]{Nitin Kumar Singh\,\orcidlink{0000-0001-6368-8437}}
\author[1,2]{Arie Rachmad Syulistyo\,\orcidlink{0000-0002-5933-1168}}
\author[1,3]{Yuichiro Tanaka\,\orcidlink{0000-0001-6974-070X}}
\author[1,3]{Hakaru Tamukoh\,\orcidlink{0000-0002-3669-1371}}

\affil[1]{Graduate School of Life Science and Systems Engineering, Kyushu Institute of Technology\\
2-4 Hibikino, Wakamatsu, Kitakyushu, 808-0196, Japan}

\affil[2]{Department of Information Technology, State Polytechnic of Malang\\
Lowokwaru, Malang, Indonesia}

\affil[3]{Research Center for Neuromorphic AI Hardware, Kyushu Institute of Technology\\
2-4 Hibikino, Wakamatsu, Kitakyushu, 808-0196, Japan}

\affil[ ]{\normalsize \textbf{Email:}\\
\texttt{nitinmjpruiitp@gmail.com}, 
\texttt{syulistyo.arie-rachmad967@mail.kyutech.jp}\\
\texttt{tanaka-yuichiro@brain.kyutech.ac.jp},
\texttt{tamukoh@brain.kyutech.ac.jp}
}
\affil[ ]{\normalsize {Copyright 2025 IEICE
, accepted by NOLTA 2025, Naha City, Okinawa, Japan, October 27--31, 2025}}
\date{}

\begin{document}
\maketitle

\begin{abstract} Sign language recognition (SLR) facilitates communication between deaf and hearing individuals. Deep learning is widely used to develop SLR-based systems; however, it is computationally intensive and requires substantial computational resources, making it unsuitable for resource-constrained devices. To address this, we propose an efficient sign language recognition system using MediaPipe and an echo state network (ESN)-based bidirectional reservoir computing (BRC) architecture. MediaPipe extracts hand joint coordinates, which serve as inputs to the ESN-based BRC architecture. The BRC processes these features in both forward and backward directions, efficiently capturing temporal dependencies. The resulting states of BRC are concatenated to form a robust representation for classification. We evaluated our method on the Word-Level American Sign Language (WLASL) video dataset, achieving a competitive accuracy of $57.71\%$ and a significantly lower training time of only $9$ seconds, in contrast to the $55$ minutes and $38$ seconds required by the deep learning-based Bi-GRU approach.
Consequently, the BRC-based SLR system is well-suited for edge devices.
\end{abstract}

%%%%%%%%%%%%%%%%%%%%%%%%%%%%%%%%%%%%%%%%%%%%%%%%%%%%%%%%%%%%%%%%%%%%%%%%%%%%%%%%
\section{Introduction}

Sign language recognition bridges the communication gap between deaf and hard-of-hearing individuals. Sign language combines various gestures, hand movements, and facial expressions to convey meaning \cite{1}. SLR technology aims to translate these gestures automatically into spoken or written language, making communication more accessible and inclusive for everyone \cite{2}. The population of individuals with hearing impairments is increasing continuously, and this trend is expected to persist in the years to come. Consequently, SLR systems are crucial in the current scenario, and researchers are seeking a reliable one that is accessible to the general public.

Deep learning-based approaches, like recurrent neural networks (RNNs) and convolutional neural networks (CNNs), are widely used by researchers for developing SLR-based systems \cite{3}. 

Deep learning-based models used for SLR have several drawbacks, including high computational demands, which make them unsuitable for edge devices such as smartphones and tablets~\cite {4,5}.

Ugale et al. review the application of CNNs in sign language recognition, also discussing the challenges of computational demands and sensitivity to individual variability \cite{ugale}.

Hossain et al. discuss the use of LSTM and 3D CNN architectures for sign language recognition. The authors also express concern over the huge training time required by LSTM and CNN, which necessitates extensive computational resources \cite{24}.

Lee et al. discuss the LSTM-RNN-based method for recognizing American Sign Language (ASL) \cite{lee2021american}. The article discusses challenges, including the need for extensive training data and potential overfitting with too many epochs.

From the above explanation, we can conclude that deep learning-based SLR systems face several challenges, especially high computational demands, which make them unsuitable for real-time use or deployment on edge devices such as smartphones, embedded systems, or tablets. As a result, optimizing these models for efficiency remains a critical area of ongoing research.

In this paper, we utilize Mediapipe to extract key features from the sign language video dataset, and these features are then fed into a bidirectional reservoir computing-based architecture for gesture classification \cite{Shherman, schaetti2018bidirectional}. The BRC improves sign language recognition by capturing gesture patterns in both forward and backward sequences. It requires minimal training, as only the output layer needs to be trained for making predictions, making it suitable for edge devices \cite{nakanishi2024bidirectional}.

\section{Materials and Methods}

\subsection{Data collection and description}

This study utilized the SLR video dataset titled WLASL 100 \cite{16}. The WLASL 100 dataset is a subset of the Word-Level American Sign Language (WLASL) dataset, which is widely used for sign language recognition-based research. The dataset comprises 1,780 training videos, 258 validation videos, and 258 testing videos, ensuring a well-balanced distribution for model training and evaluation.
WLASL 100 includes video clips performed by multiple signers, which provide variations in hand movements, speed, and execution styles for different labels, as shown in Fig.~1.

\begin{figure}[b]
  \centering
  \includegraphics[width=\linewidth]{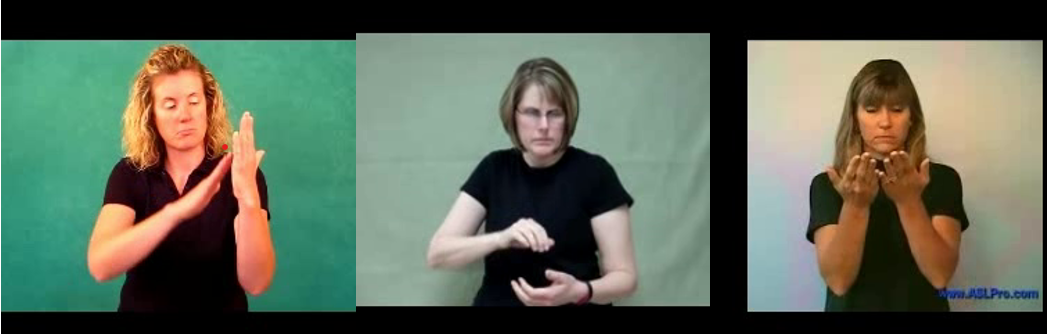}
  \caption{Signs used by different signers for activities like painting, studying, and reading.}
  \label{wlasl}
\end{figure}

\subsection{MediaPipe}

MediaPipe is a highly adaptable framework that simplifies the task of feature extraction from video datasets. It enables efficient and accurate extraction of meaningful landmarks and key points from video data \cite{33}. We can use MediaPipe’s hand, face, and pose detection modules to isolate critical visual features from each frame. These modules detect and track joints, fingertips, facial landmarks, and body poses in real-time, turning raw video data into structured, machine-readable features.

\subsection{Bidirectional reservoir computing}
In this paper, we used ESN-based BRC for SLR. The echo state network captures temporal dynamics efficiently by training only the output weights, making it fast and suitable for low-resource applications. ESN-based BRC can process the input sequence in forward and backward directions to capture past and future contexts \cite{ibrahim2022bidirectional}.
In a standard ESN, the input signal is processed only in the forward temporal direction, influencing the reservoir states sequentially over time. In contrast, a bidirectional ESN captures temporal dependencies in both forward and backward directions, thereby providing richer contextual information when dealing with sequential data.

Bidirectional Reservoir Computing (BRC) enhances the model's capability to capture temporal dependencies in tasks such as sign language recognition, speech recognition, and text analysis, where both past and future contexts can contribute to interpreting the current state.
The ESN-based BRC maps its inputs to a high-dimensional state space, and the outputs from both states are combined to predict or classify based on the complete temporal context of the input data.

State equations for bidirectional ESN-based reservoir computing are given below:

The state update equation for the forward reservoir, which processes the input sequence in the forward direction, can be defined by Equation~\ref{eq:forward_state_update}.
\begin{equation}
x_f(t+1) = (1 - \alpha)x_f(t) + \alpha\, \sigma(W_r x_f(t) + W_{\mathrm{in}} u(t))
\label{eq:forward_state_update}
\end{equation}
Where $x_f(t)$ denotes the state vector of the forward reservoir at time $t$, $u(t)$ is the input, $W_r$ is the fixed internal weight matrix of reservoir, $W_{\text{in}}$ is the input weight matrix, $\sigma$ is the activation function and $\alpha$ is the leak rate parameter.

The state update equation for the backward reservoir, which processes the time-reversed input sequence using Python-style array slicing notation {sequence[:, ::-1, :]}, can be defined by Equation~\ref{eq:backward_state_update}.

\begin{equation}
x_b(t+1) = (1 - \alpha)x_b(t) + \alpha\, \sigma\left(W_r x_b(t) + W_{\mathrm{in}} \dot{u}(t)\right)
\label{eq:backward_state_update}
\end{equation}

Where $x_b(t)$ represents the state vector of the backward reservoir at time $t$, which processes the input sequence in the backward direction, $\dot{u}(t)$ denotes the reversed input sequence, $W_r$ is the fixed internal weight matrix for the backward reservoir, $\sigma(\cdot)$ is the activation function, and $W_{\mathrm{in}}$ is the input weight matrix connecting the reversed input sequence to the backward reservoir, analogous to the forward configuration.

Here, the forward and backward processing occur sequentially. The backward direction operates on the time-reversed input sequence. The resulting forward and backward states are then concatenated, as expressed in Equation~\ref{eq:combined_output}, to produce the final output:
\begin{equation}
y(t) = W_{\text{out}} (x_f(t) \oplus x_b(t)) \label{eq:combined_output}
\end{equation}where $y(t)$ is the output vector at time $t$, $W_{\text{out}}$ is the trained output weight matrix that maps the concatenated states from both the forward and backward reservoirs to the target output, and $\oplus$ denotes the concatenation of both states of BRC. We used ridge regression to train $W_{\text{out}}$ and classify the labels.

In this paper, we use the activation function $\sigma(\cdot) = \tanh(\cdot)$.
To improve the model's flexibility, a bias term or a small noise component can be incorporated into the reservoir dynamics. The leak rate, $\alpha \in [0, 1]$, controls the speed at which the reservoir state updates. For bidirectional processing, both the forward and backward reservoir states are computed using the same reservoir configuration, sharing identical input weight matrix $W_{\mathrm{in}}$ and fixed internal weight matrix $W_r$.

\subsection{Ridge regression}

We used ridge regression to train and optimize the bidirectional reservoir computing model for SLR. Ridge regression is a form of linear regression incorporating a regularization technique to prevent overfitting and improve the model’s generalization to unseen data \cite{25}.

\subsection{Bidirectional gated re-current unit (Bi-GRU) }
We also employed the deep learning-based method Bi-GRU for SLR and compared the results with a BRC-based architecture. We fed the keypoints extracted from the WLASL100 video dataset (by using MediaPipe) to the Bi-GRU-based architecture.

\section{Results and discussion}

This article compares SLR on the WLASL 100 video dataset using bidirectional ESN-based RC (with 100 nodes in both cases, i.e., forward and backward), Bi-GRU (with 150 epochs), and single standard ESN-based RC (with unidirectional RC and 200 nodes). We used MediaPipe to extract key points from the WLASL 100 dataset in all methods. All experiments were performed on intel\textsuperscript{\textregistered} core\textsuperscript{™} i7-11700 processor.
We utilized MediaPipe to extract key features from the WLASL100 video dataset, as illustrated in Fig.~2. MediaPipe offers real-time, robust hand tracking and landmark detection, which is essential for capturing the nuanced gestures present in the sign language video dataset. By extracting precise coordinates of hand joints frame by frame, MediaPipe enables our model to accurately capture the dynamic movements characteristic of sign language, providing a rich set of spatial and temporal features.

  \begin{figure} [b]
    \centering
    \includegraphics[width=\linewidth]{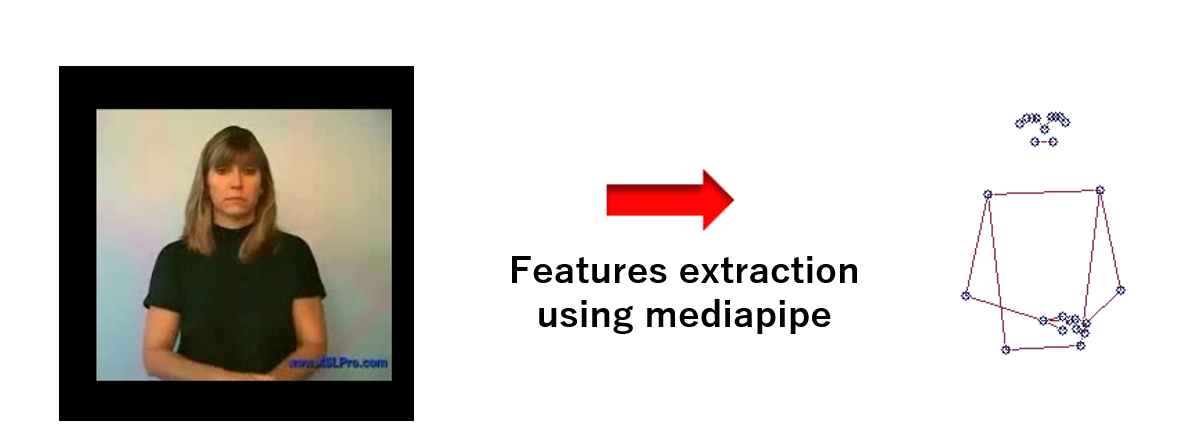}
    \caption{Feature extraction using MediaPipe}
    \label{fig:zcr}
\end{figure}

Fig.~3 illustrates SLR using a bi-directional reservoir computing-based architecture. As explained in detail above, the MediaPipe framework is used to extract key points from each video frame. 

\begin{figure}[t]
    \centering
    \includegraphics[width=\linewidth]{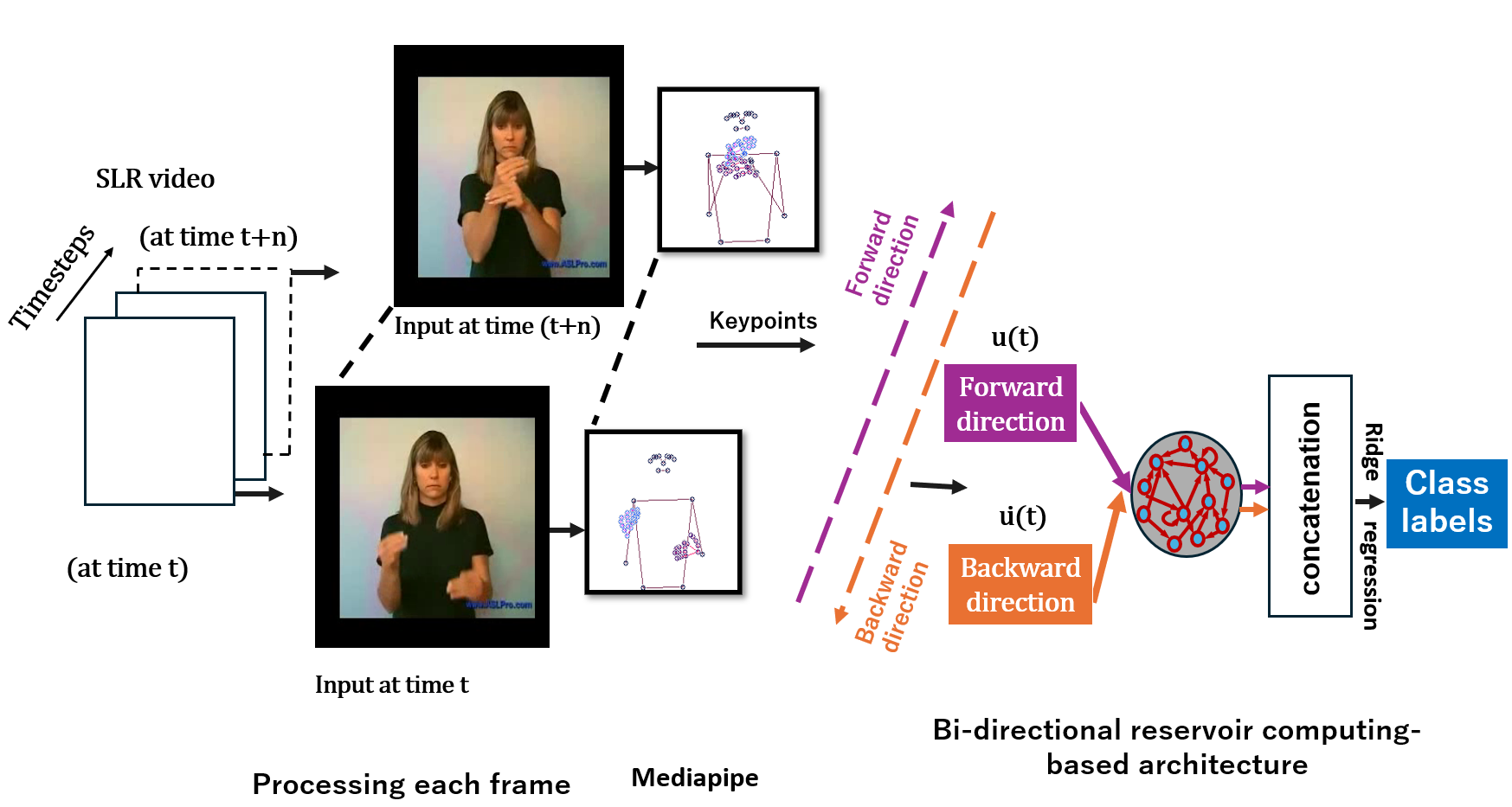}
    \caption{SLR system using Bidirectional reservoir computing}
    \label{light gbm}
\end{figure}
These keypoints are further processed using a bidirectional reservoir computing architecture having a total of 200 nodes after concatenation. This architecture processes keypoints in both forward and backward directions: the forward direction captures the progressive dynamics of hand gestures, while the backward direction analyzes them retrospectively. Both directions are essential for capturing the full temporal dynamics of the gestures, with the outputs concatenated to form a comprehensive feature set. Ridge regression is then used to map these features to the final output labels, enabling classification of signs from the WLASL 100 dataset.

This architecture leverages MediaPipe's strengths for precise spatial feature extraction and bidirectional reservoir computing's dynamic temporal processing capabilities to enhance the performance of sign language recognition. Additionally, only the output layer needs to be trained in the reservoir, thereby reducing training time.

Table 1 compares the performance of three sign language recognition methods implemented on the WLASL100 dataset: the proposed bidirectional ESN reservoir computing approach, single unidirectional ESN, and Bi-GRU. We apply these machine learning algorithms to the WLASL100 video dataset. In all the compared methods, we extracted keypoints from the WLASL100 video dataset using MediaPipe and fed them into the corresponding architectures (i.e., standard single ESN, Bidirectional ESN, and Bi-GRU) for classifying sign language data. The metrics shown in Table 1 include accuracy (with standard deviation) and training time, highlighting the trade-offs between computational efficiency and recognition performance.

\begin{table}[t]
\caption{SLR on WLASL100 dataset}
\centering
\begin{tabular}{lcc}
\toprule
\textbf{Methods used for SLR} & \textbf{Accuracy (\%) $\pm$ SD} & \textbf{Training time (min:sec)} \\ 
\midrule
Proposed method (bi-directional ESN)      & 57.71 $\pm$ 1.35 & 00:09 \\
Standard single ESN (uni-directional)     & 54.31 $\pm$ 1.45 & 00:07 \\
Bi-GRU                                     & 49.90 $\pm$ 2.56 & 55:38 \\
\bottomrule
\end{tabular}
\label{table:slr_methods}
\end{table}

The proposed bidirectional reservoir computing approach achieves the highest accuracy ($57.71\% \pm 1.35$) with a training duration of only $9$ seconds, significantly lower than the 55 minutes and 38 seconds required by the deep learning-based Bi-GRU method. The results shown in Table 1 show that the SLR-based BRC system is well-suited for real-time applications on edge devices.

\section{Conclusion}

This paper compares a bidirectional ESN-based architecture with a Bi-GRU and a standard single unidirectional ESN-based RC system for SLR. The results presented in this paper demonstrate that the proposed BRC approach achieves the competitive accuracy of $57.71\%$ with a low training time of only 9 seconds, which is significantly lower than that of the deep learning-based Bi-GRU model. These results conclude that the proposed ESN-based BRC architecture, combined with MediaPipe, offers a promising solution for deploying SLR systems on resource-constrained devices. 


%%%%%%%%%%%%%%%%%%%%%%%%%%%%%%%%%%%%%%%%%%%%%%%%%%%%%%%%%%%%%%%%%%%%%%%%%%%%%%%%

%%%%%%%%%%%%%%%%%%%%%%%%%%%%%%%%%%%%%%%%%%%%%%%%%%%%%%%%%%%%%%%%%%%%%%%%%%%%%%%%

%%%%%%%%%%%%%%%%%%%%%%%%%%%%%%%%%%%%%%%%%%%%%%%%%%%%%%%%%%%%%%%%%%%%%%%%%%%%%%%%

%%%%%%%%%%%%%%%%%%%%%%%%%%%%%%%%%%%%%%%%%%%%%%%%%%%%%%%%%%%%%%%%%%%%%%%%%%%%%%%%

\section*{Acknowledgement}
This work is based on results obtained from a project JPNP16007, commissioned by the New Energy and Industrial Technology Development Organization (NEDO) and supported by JST ALCA-Next Grant Number JPMJAN23F3 and JSPS KAKENHI Grant numbers 22K17968, 23H03468, and 23K18495.

\bibliography{ref}

@article{1,
  title={Global hearing health care: new findings and perspectives},
  author={Wilson, Blake S and Tucci, Debara L and Merson, Michael H and O'Donoghue, Gerard M},
  journal={The Lancet},
  volume={390},
  number={10111},
  pages={2503--2515},
  year={2017},
  publisher={Elsevier}


}

@article{2,
  title={Sign language recognition systems: A decade systematic literature review},
  author={Wadhawan, Ankita and Kumar, Parteek},
  journal={Archives of computational methods in engineering},
  volume={28},
  pages={785--813},
  year={2021},
  publisher={Springer}
}

@article{3,
  title={Static sign language recognition using deep learning},
  author={Tolentino, Lean Karlo S and Juan, RO Serfa and Thio-ac, August C and Pamahoy, Maria Abigail B and Forteza, Joni Rose R and Garcia, Xavier Jet O},
  journal={International Journal of Machine Learning and Computing},
  volume={9},
  number={6},
  pages={821--827},
  year={2019},
  publisher={EJournal Publishing}
}

@article{4,
  title={A comprehensive study on deep learning-based methods for sign language recognition},
  author={Adaloglou, Nikolas and Chatzis, Theocharis and Papastratis, Ilias and Stergioulas, Andreas and Papadopoulos, Georgios Th and Zacharopoulou, Vassia and Xydopoulos, George J and Atzakas, Klimnis and Papazachariou, Dimitris and Daras, Petros},
  journal={IEEE transactions on multimedia},
  volume={24},
  pages={1750--1762},
  year={2021},
  publisher={IEEE}
}

@article{5,
  title={Low-cost computation for isolated sign language video recognition with multiple reservoir computing},
  author={Arie Rachmad Syulistyo and N Fuengfusin and Yuichiro Tanaka and Hakaru Tamukoh},
  journal={PLOS ONE, accepted},
  year={2025}
}

@inproceedings{ugale,
  title={A Review on Sign Language Recognition Using CNN},
  author={Ugale, Meena and Shinde, Odrin Rodrigues Anushka and Desle, Kaustubh and Yadav, Shivam},
  booktitle={International Conference on Applications of Machine Intelligence and Data Analytics (ICAMIDA 2022)},
  pages={251--259},
  year={2023},
  organization={Atlantis Press}
}

@article{lee2021american,
  title={American sign language recognition and training method with recurrent neural network},
  author={Lee, Carman KM and Ng, Kam KH and Chen, Chun-Hsien and Lau, Henry CW and Chung, Sui Ying and Tsoi, Tiffany},
  journal={Expert Systems with Applications},
  volume={167},
  pages={114403},
  year={2021},
  publisher={Elsevier}
}

@article{Shherman,
  title={Method Development Through Landmark Point Extraction for Gesture Classification With Computer Vision and MediaPipe.},
  author={Suherman, Suherman and Suhendra, Adang and Ernastuti, Ernastuti},
  journal={TEM Journal},
  volume={12},
  number={3},
  year={2023}
}

@inproceedings{16,
  title={Word-level deep sign language recognition from video: A new large-scale dataset and methods comparison},
  author={Li, Dongxu and Rodriguez, Cristian and Yu, Xin and Li, Hongdong},
  booktitle={Proceedings of the IEEE/CVF winter conference on applications of computer vision},
  pages={1459--1469},
  year={2020}
}

@article{schaetti2018bidirectional,
  title={Bidirectional Echo State Network-based Reservoir Computing for Cross-domain Authorship Attribution},
  author={Schaetti, Nils},
  journal={Notebook for PAN at CLEF},
  year={2018}
}

@article{nakanishi2024bidirectional,
  title={Bidirectional 2D reservoir computing for image anomaly detection without any training},
  author={Nakanishi, Keiichi and Tokunaga, Terumasa},
  journal={Nonlinear Theory and Its Applications, IEICE},
  volume={15},
  number={4},
  pages={838--850},
  year={2024},
  publisher={The Institute of Electronics, Information and Communication Engineers}
}

@article{ibrahim2022bidirectional,
  title={Bidirectional parallel echo state network for speech emotion recognition},
  author={Ibrahim, Hemin and Loo, Chu Kiong and Alnajjar, Fady},
  journal={Neural Computing and Applications},
  volume={34},
  number={20},
  pages={17581--17599},
  year={2022},
  publisher={Springer}
}

@inproceedings{24,
  title={Sign language recognition analysis using multimodal data},
  author={Santhalingam, Panneer Selvam and Pathak, Parth and Ko{\v{s}}eck{\'a}, Jana and Rangwala, Huzefa and others},
  booktitle={2019 IEEE International Conference on Data Science and Advanced Analytics (DSAA)},
  pages={203--210},
  year={2019},
  organization={IEEE}
}

@inproceedings{25,
  title={Stable output feedback in reservoir computing using ridge regression},
  author={Wyffels, Francis and Schrauwen, Benjamin and Stroobandt, Dirk},
  booktitle={International conference on artificial neural networks},
  pages={808--817},
  year={2008},
  organization={Springer}
}

@inproceedings{33,
  title={Preprocessing Mediapipe Keypoints with Keypoint Reconstruction and Anchors for Isolated Sign Language Recognition},
  author={Roh, Kyunggeun and Lee, Huije and Hwang, Eui Jun and Cho, Sukmin and Park, Jong C},
  booktitle={Proceedings of the LREC-COLING 2024 11th Workshop on the Representation and Processing of Sign Languages: Evaluation of Sign Language Resources},
  pages={323--334},
  year={2024}
}
\bibliographystyle{IEEEtran}

\appendix
\section*{Appendix}

\begin{itemize}
    \item This work has been accepted by the International Symposium on Nonlinear Theory and Its Applications (NOLTA-2025), Naha City, Okinawa, Japan, October 27--31, 2025. \url{https://nolta2025.org/}
    \item The video abstract related to this work is available at:\\
    \url{https://www.youtube.com/watch?v=WLdyJ-aK-mo}
\end{itemize}

\end{document}